# Human and Smart Machine Co-Learning with Brain Computer Interface

*Chang-Shing Lee, Mei-Hui Wang, Li-Wei Ko, Naoyuki Kubota*
*Lu-An Lin, Shinya Kitaoka, Yu-Te Wang, and Shun-Feng Su*

## Abstract

Machine learning has become a very popular approach for cybernetics systems, and it has always been considered important research in the Computational Intelligence area. Nevertheless, when it comes to smart machines, it is not just about the methodologies. We need to consider systems and cybernetics as well as include human in the loop. The purpose of this article is as follows: (1) To integrate the open source Facebook AI Research (FAIR) DarkForest program of Facebook with Item Response Theory (IRT), to the new open learning system, namely, DDF learning system; (2) To integrate DDF Go with Robot namely Robotic DDF Go system; (3) To invite the professional Go players to attend the activity to play Go games on site with a smart machine. The research team will apply this technology to education, such as, playing games to enhance the children concentration on learning mathematics, languages, and other topics. With the detected brainwaves, the robot will be able to speak some words that are very much to the point for the students and to assist the teachers in classroom in the future.

## I.  Introduction

Rémi Coulom, a freelance developer of Go programs, said "*Online games are usually played at a faster pace, which favors the computer over humans,*" and he still expected a strong correlation with performance in serious tournament games [1]. Hence, the held special event *Human and Smart Machine Co-Learning @ IEEE SMC 2017* (http://oase.nutn.edu.tw/IEEESMC2017/, Banff, Canada, Oct. 5, 2017) still has Go games on site, not just playing Go on the Internet. The purpose of the activity in IEEE SMC 2017 is as follows: 1) To integrate the open source Facebook AI Research (FAIR) DarkForest (DF) program of Facebook (USA) [2] with Item Response Theory (IRT) [8] of NUTN, Taiwan, to the new open learning system, namely, Dynamic DF (DDF, Dynamic DarkForest) learning system [3]; 2) To integrate DDF Go with FUJISOFT Robot (led by Kubota Lab., TMU, Japan) namely Robotic DDF Go system; 3) To invite the professional Go players to attend the activity to play Go games on site with a smart machine. Chun-Hsun Chou (9P, Taiwan), Ping-Chiang Chou (6P, Taiwan), and Kai-Hsin Chang (5P, Taiwan) were invited to play Go games with *DeepZenGo* (Japan). Lu-An Lin (6D, Taiwan), Daisuke Horie (4D, Japan), and Shuji Takemura (1D, Japan) played games with *Dynamic Darkforest* (*DDF*, Taiwan) embedded FAIR Darkforest Open Go AI Engine [2]. In addition, the research collaborative team from National Chiao Tung University (NCTU), National University of Tainan (NUTN), University of California San Diego (UCSD), and National Center for High-Performance Computing (NCHC) jointly integrated the Brain Computer Interface (BCI) with the current Dynamic-Darkforest (called BCI-DDF) Go system, which was also firstly demonstrated, to attract more scholars in the brain machine interaction (BMI) area in IEEE SMC conference and then join the SMC society.

## II.  Past held events in the world from 2008 to 2017

Owing to the maturity of deep learning technologies and computer hardware, Google combined them together with Monte Carlo Tree to beat many top professional Go players without handicaps in 2016 and 2017 [4-5]. This year is the first year to hold Human & Smart Machines Co-Learning @ IEEE SMC 2017. However, we have carried out the events of humans playing Go with the computer Go programs for almost a decade [6-7]. Fig. 1 shows the past held events of Human vs. Computer Go Competitions from 2008 to 2017 (https://www.youtube.com/watch?v=UkSOVnbC2Y8) funded by IEEE CIS, IEEE SMC, Taiwanese government, NUTN, and Taiwanese Association for Artificial Intelligence (TAAI). In 1998, the handicaps for the human vs. computer 19×19 game were 29 stones [6]. However, the power of computer Go programs has increased from seven-stone to zero-stone handicap against top professional Go players from 2008 to 2017.

## III.  BCI-DDF Go System

In the special event of IEEE SMC 2017, we combined the theory of deep learning with the technology of BCI [9, 10] to demonstrate playing Go. The brainwave technology has been developed for a long time; however, applying it to play Go is the world's first case in an IEEE conference. The world latest mobile and wireless EEG system is fully utilized in the innovation of the developed BCI-DDF Go system. The wireless system, developed by the research team from Brain Research Center in NCTU, is designed to extract the Go player's brainwaves when they play and compete with the DDF Go system directly. Fig. 2 shows the two-mode (a *competitive learning* mode and a *predictive learning* mode) scenario of *Human & Smart Machine Co-Learning @ IEEE SMC 2017*, including the invited Go players, computer Go programs, robot Palro, and the developed BCI-DDF Go system. Fig. 3 shows the BCI-DDF Go system diagram that Lu-An Lin (6D) played Go with DDF without using her hand and the robot Palro reported her the next moves suggested by DDF, which were demonstrated in IEEE SMC 2017.

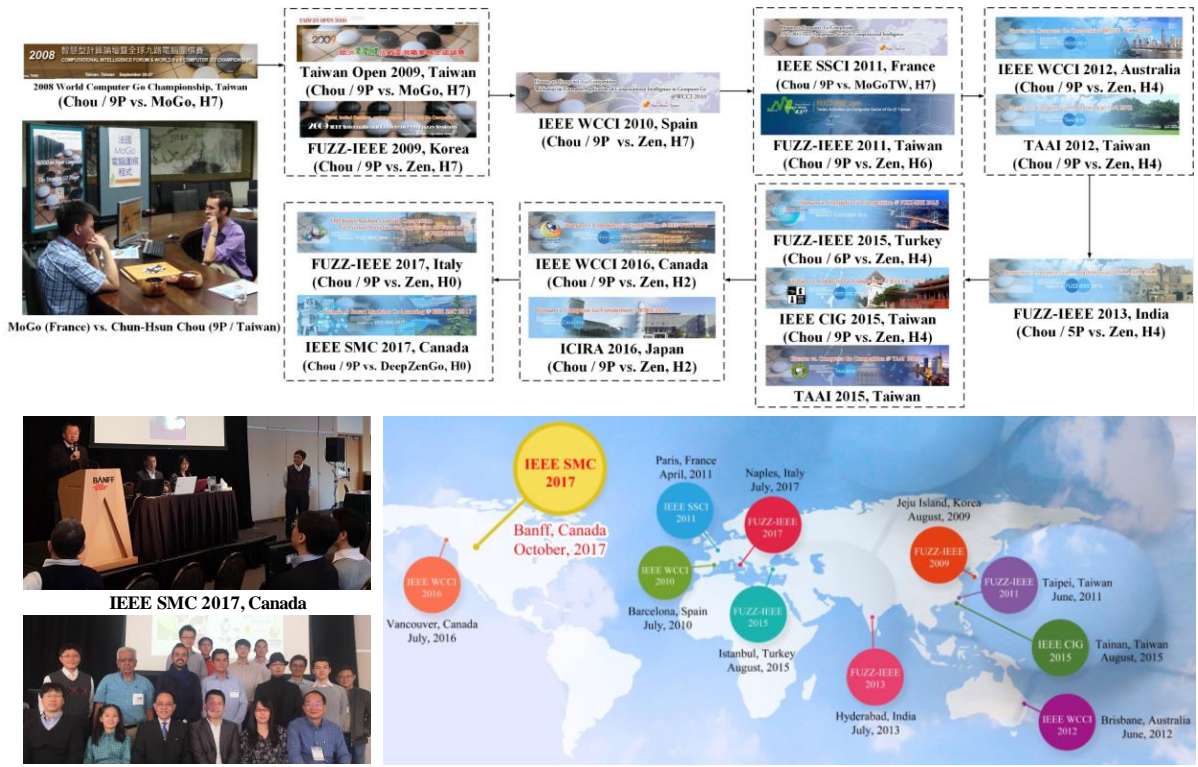

Figure 1.    Past held events in the world from 2008 to 2017.

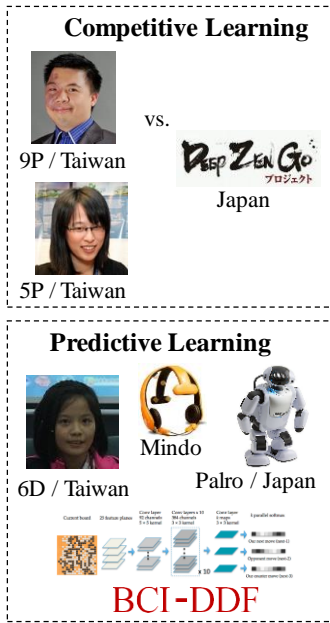

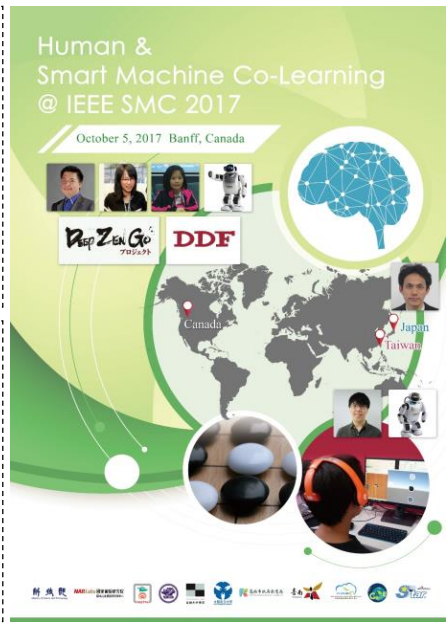

Figure 2.    Two-mode scenario of Human & Smart Machine Co-Learning @ IEEE SMC 2017, including the invited Go players, computer Go programs, robot Palro, and the developed BCI-DDF Go system.

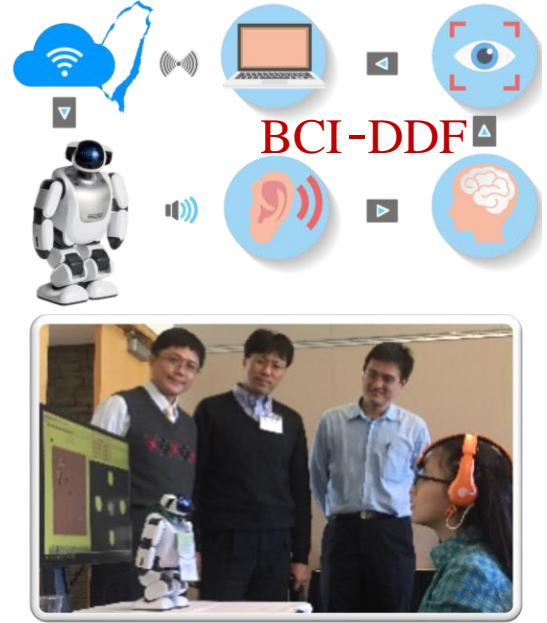

Figure 3.    BCI-DDF Go system diagram that Lu-An Lin (6D, right 1) played Go with DDF without using her hand, and the robot Palro reported her the next moves suggested by DDF.

We adopted the steady-state visual evoked potential (SSVEP) technology to collect the brain signals from the visual cortex (O1 and O2 channels) and performed the real-time signal processing in the cloud server. Five pilot players have tested the developed BCI-DDF Go system and it can reach about 90% accuracy on a five-class classification task before we held this special event. The demonstration of the BCI-DDF Go system is expressive of the breakthrough in human brain and interaction with the artificial intelligence. Humans wore a wireless EEG headset and were instructed to gaze at the coded visual stimulus on the screen. The BCI-DDF GO system continuously decoded the on-going EEG to send the move command to the game server. While the Go player is wearing the wireless EEG headset, the technician must check the impedance between the scalp and the EEG electrodes first for collecting the good quality of EEG signals.

Fig. 4(a) shows the impedance map of O1 and O2 channels (~100 KOhm). Fig. 4(b) shows the Go player's EEG signals collected from O1 and O2 channels in which he / she was successfully perceiving the 8Hz visual stimulus for controlling the up Arrow. Then, the DDF communicates with BCI via the WebSocket protocol. The decoded EEGs are sent to the NCHC server in Taiwan, and then to FAIR Darkforest AI Go Engine to control the stone to move. Humans do not need to use their hands to play Go anymore, and even they can play Go with the robot together because the robot is able to predict and report the next suggested moves to them for a reference before playing next move. Epoch Times Calgary also reported this special event by the topic of "*The world's first playing Go using brainwaves in Banff*" in 2017 (https://www.youtube.com/watch?v=okMA0Snhj-s).

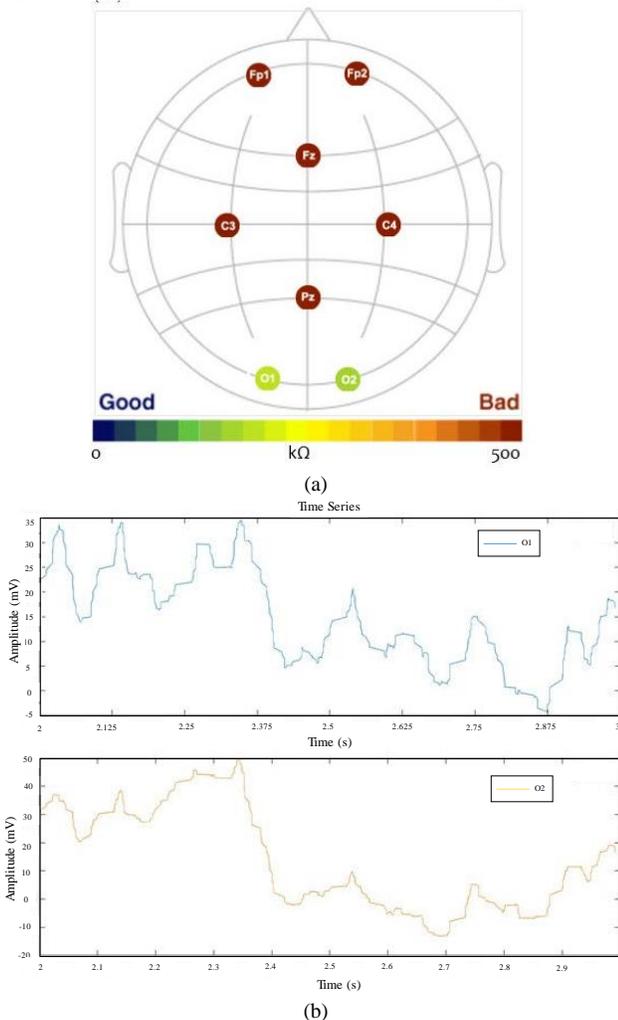

Figure 4.    (a) Impedance check of the EEG channels and (b) Go player's EEG signals extracted from O1 and O2 channels for controlling the up arrow of the BCI-DDF Go system.

## IV.  Game results

One of the top computer Go programs, *DeepZenGo*, from Japan was used to play with the invited professional Go players, including Chun-Hsun Chou (9P), Ping-Chiang Chou (6P), and Kai-Hsin Chang (5P) for seven games without handicaps. This mode is a learning mode between humans and machines. In order to make *DeepZenGo* have different strength while it plays with Chou (9P) and Chou (6P) / Chang (5P), *DeepZenGo* runs (1 node / E5-2643 v4 x2 / NVIDIA Titan X (Pascal) RAM 128GB / Storage SSD480GB) and (1 node / E5-2623 v3 x2 / NVIDIA Titan X RAM 128GB / Storage SSD480GB), respectively. *DeepZenGo* won all of the games with three invited professional Go players. Figs. 5(a)-(b) are the games where Chou (9P) as White / *DeepZenGo* as Black (B + Resign), as well as Chang (5P) as White / *DeepZenGo* as Black (B + Resign), respectively.

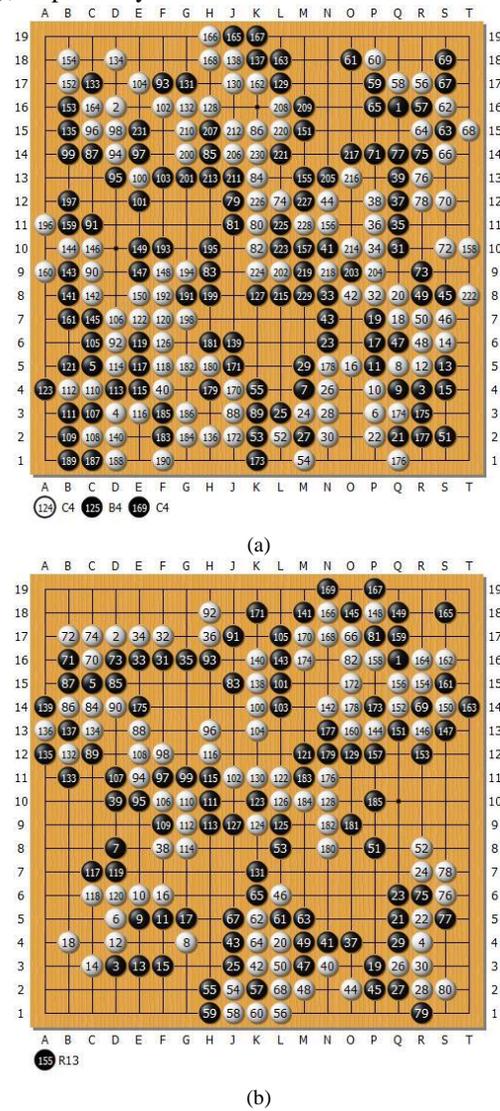

Figure 5.    (a) Chou (9P) as White, *DeepZenGo* as Black, and Black won the game with resignation and (b) Chang (5P) as White, *DeepZenGo* as Black, and Black won the game with resignation.2017.

Chou (9P) talked of his experience after the event: "*This is my first experience to play Go games with DeepZenGo in public and I am obviously at a disadvantage after the layout of the three games. The rise of AI challenges professional Go players' domain knowledge of the game layout which has been learned and taught for hundreds of years. I have joined the human vs. computer Go competitions hosted by NUTN and other academic organizations since 2008. In the past, I advised computer Go programs; however, I am advised now instead.*"

Chang (5P) said "*It is an indisputable fact that top-level*

*computer Go programs, like AlphaGo and DeepZenGo, can completely defeat the professional Go players. I hadn't yet personally experienced the strength of the artificial intelligence. But, I played with DeepZenGo in the special event of IEEE SMC 2017. In the past, the layout and the situation judgment of Go were two of the most difficult problems to solve for computer Go programs; however, now, the top-level machine does much better than humans, not to mention its ability to handle life and death. AI overturns domain knowledge that humans have been learned and taught. In a sense, humans are given a different picture of the world. AI has simulated the rich variety of playing-Go models and led Go to another level.*"

Lin (6D) was invited to play Go with DDF by brainwaves and cooperated with the robot Palro under the predictive learning mode. Take the game (Lin + Palro as Black, DDF as White) for example. Fig. 6(a) shows the curves of the predictive winning rate and numbers of simulations. Fig. 6(b) shows the inferred game-situation results of the developed FML-based dynamic assessment agent, where $B^{++}$, $B^+$, $U$, $W^+$, and $W^{++}$ represent "*Black is obvious at an advantage,*" "*Black is possible at an advantage,*" "*Both are in an uncertain situation,*" "*White is possible at an advantage,*" and "*White is obvious at an advantage,*" respectively [3]. After the event, Lin said that "*This is my first time to attend the international conference. Indeed, it is such an amazing experience, in which wearing a Mindo to play Go was a real eye-opener for me. When the sensors in the headgear measured my brainwaves continually, the sparkling points on the monitor caused my eyes a little bit of discomfort. But, this did not dampen my interest in this biometric gadget and enthusiasm for playing Go using brainwaves. This developed technology can be applied in a wide variety of fields; for example, the physically challenged can use it to express their inner thoughts. Glad to have this great opportunity to wear this biometric headgear to play Go and hope to see its further development, https://www.youtube.com/watch?v=6SuZ92z_PBU&feature=youtu.be).*"

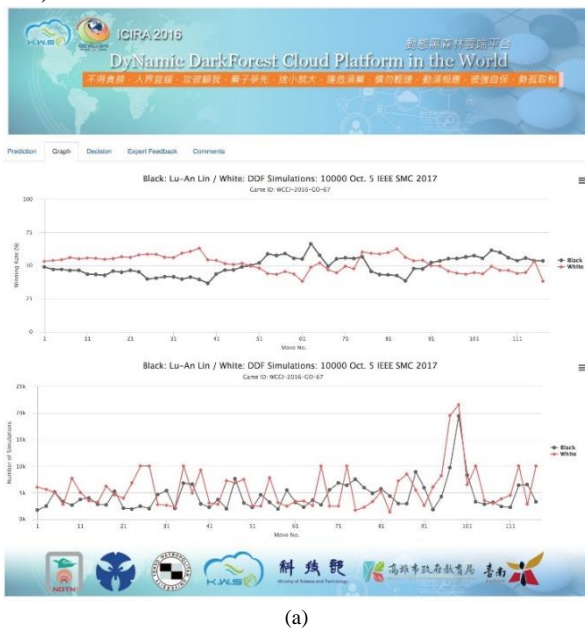

(a)

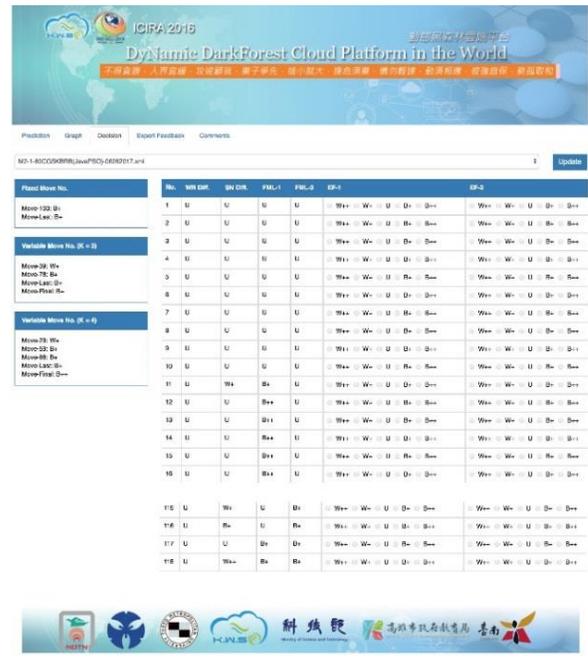

(b)

Figure 6.    (a) Predictive winning rate and numbers of simulations curves and (b) inferred game-situation results of FML-based dynamic assessment agent for the game that (Lin + Palro) as Black, DDF as White, and Black won the game.

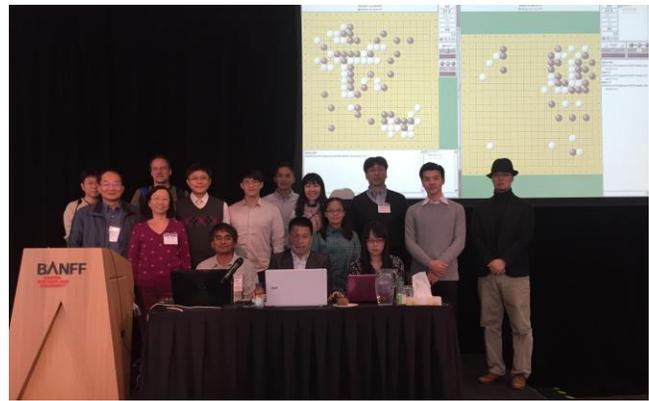

Figure 7.    Special event venue, where General Chair (Anup Basu, foreground, left 1), Program Chair (Irene Cheng, the left of General Chair), Yo-Ping Huang (the left of Program Chair), and Shinya Kitaoka (background, right 1).

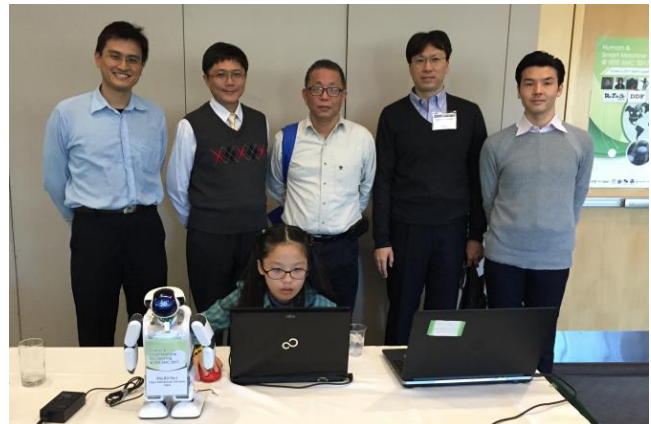

Figure 8.    Li-Wei Ko, Chang-Shing Lee, Shun-Feng Su, Naoyuki Kubota, Takenori Obo (background, from left to right), Lu-An Lin (6D, foreground) playing Go with DDF, and the robot Palro reported real-time suggested next move to Lin.

## V. Conclusion

Smart machine is one of the main themes of IEEE SMC 2017. A study on Go players competing with and learning from smart machine is sure to attract more attention from worldwide scholars in SMC conferences. There are 15 games played at this special event where the amateur Go players won 8 games against DDF with the small computational resource, but *DeepZenGo* won all the 7 games against the invited professional Go players. Therefore, *DeepZenGo* could be the coach of the professional Go players while DDF with small machine could be the student of the amateur Go players as long as we properly and adaptively adjust the computational resources of the intelligent robot. Figs. 7 and 8 show some of the selected photos for this special event. In the future, the research team will apply this technology to education, such as, playing games to enhance children's concentration on learning mathematics, languages, and other topics. With the detected brainwaves, the robot will be able to speak some words that are very much to the point for the students and to assist the teachers in class in the future.

## Acknowledgement

The event in IEEE SMC 2017 was a great success and we would like to express our heartfelt thanks to everyone who offered any help, joined, as well as watched the games and the BCI-DDF demonstration. We would also like to sincerely thank the IEEE SMC Society, IEEE SMC President Dimitar Filev, Sr. IEEE SMC Past President Philip Chen, the organizing committee of the IEEE SMC 2017, especially General Chair Anup Basu. The authors would like to thank the Ministry of Science and Technology of Taiwan (MOST 106-3114-E-024-001) and IEEE SMC's outreach project for their financial support. Finally, we would like to thank three professional Go players (Chun-Hsun Chou, Kai-Hsin Chang, and Ping-Chiang Chou), Sheng-Chi Yang, Taiwanese government, NCHC of Taiwan, and Osaka Prefecture University (OPU / Japan) for supporting this special event.

## About the authors

*Chang-Shing Lee* (leecs@mail.nutn.edu.tw) received the Ph.D. degree in Computer Science and Information Engineering from the National Cheng Kung University, Tainan, Taiwan, in 1998. He is currently a Professor with the Department of Computer Science and Information Engineering, National University of Tainan. He was awarded Certificate of Appreciation for outstanding contributions to the development of IEEE Standard 1855TM-2016 (IEEE Standard for Fuzzy Markup Language). He was the Emergent Technologies Technical Committee (ETTC) Chair of the IEEE Computational Intelligence Society (CIS) from 2009 to 2010 and serves as the AE of IEEE TCIAIG.

*Mei-Hui Wang* (mh.alice.wang@gmail.com) received M.S. degree in electrical engineering from the Yuan Ze University, Chung-Li, in 1995. She is currently a Researcher with the Ontology Application and Software Engineering (OASE) Laboratory, Department of Computer Science and Information Engineering, National University of Tainan (NUTN), Taiwan.

*Li-Wei Ko* (lwko@mail.nctu.edu.tw) received the Ph.D. degree in electrical engineering from National Chiao Tung University, Hsinchu, Taiwan in 2007. Currently, he is an associate professor with the Institute of Bioinformatics and Systems Biology, National Chiao Tung University, Taiwan. Dr. Ko is an AE of IEEE Transactions on Neural Networks and Learning Systems.

*Naoyuki Kubota* (kubota@tmu.ac.jp) received the D. E. degree from Nagoya University, Nagoya, Japan in 1997. Currently, he is a professor in the Department of System Design, Tokyo Metropolitan University, Tokyo, Japan.

*Lu-An Lin* (luan20050427@gmail.com), a 12-year-old 7th grader in Taiwan, is a 6D amateur Go player. Ever since she was in 6th grade, she has been cooperating with OASE Lab., NUTN, Taiwan to carry out research into smart machine learning of Go and language.

*Shinya Kitaoka* (shinya_kitaoka@dwango.co.jp) is one of the DeepZenGo team members and currently, he is an engineer with Dwango Media Village, DWANGO Co., Ltd., Japan.

*Yu-Te Wang* (yutewang65@gmail.com) received the Ph.D. degree in computer science and engineering at the University of California San Diego (UCSD), La Jolla, in 2015. Currently, he is a Staff Research Associate III at the Swartz Center for Computational Neuroscience, UCSD, USA.

*Shun-Feng Su* (su@orion.ee.ntust.edu.tw) received the Ph.D. degrees in electrical engineering from Purdue University, West Lafayette, IN, USA, in 1991. He is currently a Chair Professor with the Department of Electrical Engineering, National Taiwan University of Science and Technology, Taiwan. Dr. Su currently serves as an AE for the IEEE Transactions on Cybernetics, the IEEE Transactions on Fuzzy Systems, and the IEEE Access, a Subject Editor (Electrical Engineering) for the Journal of the Chinese Institute of Engineers, and the Editor-in-Chief for the International Journal of Fuzzy Systems.